\documentclass{article}
\usepackage{spconf,amsmath,graphicx,hyperref}
\usepackage{graphicx}   
\usepackage{subcaption} 
\usepackage{amsmath} 
\usepackage{caption}
\usepackage{bm}      
\usepackage{enumitem}
\usepackage{booktabs} 
\usepackage{xcolor}   

\usepackage{colortbl} 
\usepackage{amsmath}  
\usepackage{amsmath, bm}
\usepackage{empheq}
\usepackage{bm}
\usepackage{graphicx} 
\usepackage{caption}
\usepackage{graphicx}

\usepackage{lipsum}
\usepackage{booktabs}   
\usepackage{tabularx}   
\usepackage{makecell}   
\usepackage{amssymb}    
\usepackage{amsmath}    
\usepackage{multirow}   
\usepackage{array}      
\usepackage{cite}

\newcolumntype{C}{>{\centering\arraybackslash}X}

\title{From Knowing to Doing Precisely: A General Self-Correction and Termination Framework for VLA models}
\name{Wentao Zhang$^{\star\dagger}$, Aolan Sun$^{\dagger\ddagger}$,  Wentao Mo$^{\star\dagger}$, Xiaoyang Qu$^{\dagger}$, Yuxin Zheng$^{\star\dagger}$, Jianzong Wang$^{\dagger\ddagger}$
\thanks{This work is supported by Shenzhen-Hong Kong Joint Funding Project (Category A) under grant No. SGDX20240115103359001. \\
\indent $^{\ddagger}$Corresponding authors are Aolan Sun (aolan\_sun@163.com) and Jianzong Wang  (jzwang@188.com).}}
\address{$^{\star}$Tsinghua Shenzhen International Graduate School, Tsinghua University, Shenzhen, China,\\
        $^{\dagger}$Ping An Technology (Shenzhen) Co., Ltd., Shenzhen, China}

\begin{document}
\ninept
\maketitle
\begin{abstract}
While vision–language–action (VLA) models for embodied agents integrate perception, reasoning, and control, they remain constrained by two critical weaknesses: first, during grasping tasks, the action tokens generated by the language model often exhibit subtle spatial deviations from the target object, resulting in grasp failures; second, they lack the ability to reliably recognize task completion, which leads to redundant actions and frequent timeout errors. To address these challenges and enhance robustness, we propose a lightweight, training-free framework—\textbf{VLA-SCT}. This framework operates as a self-correcting control loop, combining data-driven action refinement with conditional logic for termination. Consequently, compared to baseline approaches, our method achieves consistent improvements across all datasets in the LIBERO benchmark, significantly increasing the success rate of fine manipulation tasks and ensuring accurate task completion, thereby promoting the deployment of more reliable VLA agents in complex, unstructured environments.
\end{abstract}
\vspace{-0.2cm}
\section{Introduction}

Vision-Language-Action (VLA) models represent a significant advancement in embodied artificial intelligence  \cite{gu2025humanoid,liurdt}. These models achieve this by integrating visual perception, natural language processing, and action generation within a unified architectural framework, which enables robotic agents to interpret high-level human instructions and execute corresponding physical tasks with considerable adaptability \cite{11127486}. Leveraging sophisticated semantic understanding, VLA models exhibit robust generalization capabilities, particularly in navigating complex environments and manipulating novel objects. Pioneering implementations in the field of VLA, such as Google DeepMind's RT-1 \cite{brohan2022rt} and RT-2 \cite{zitkovich2023rt}, demonstrated that joint training on large-scale internet data alongside robotic trajectory datasets endows these models with the capacity for generalization to unseen objects and novel commands. Subsequent advancements in this domain include the development of the RT-X series \cite{o2024rt-x}, the open-source model OpenVLA \cite{kim2025openvla}, and the more recent $\pi$ series, including $\pi_{0}$
  \cite{black2410pi0} and $\pi_{0.5}$
  \cite{black2025pi0.5}, which leverage flow matching strategies. To improve the performance of these models, researchers have adopted techniques such as model pruning and token sparsification to accelerate inference, such as SparseVLM\cite{zhangsparsevlm} and FastVLM\cite{vasu2025fastvlm}.

Although these methods can improve inference speed to a certain extent, they greatly reduce the success rate of the model in performing tasks, which hinders their reliable deployment in real-world scenarios\cite{zhong2025survey}. The key problem in the current VLA models is the knowledge-action gap. First, a primary reflection of this gap is the models' lack of precision in fine-grained manipulation tasks. This  
\begin{figure}[t]
    \centering 

    \begin{subfigure}{\linewidth} 
        \centering
        \includegraphics[width=0.9\linewidth]{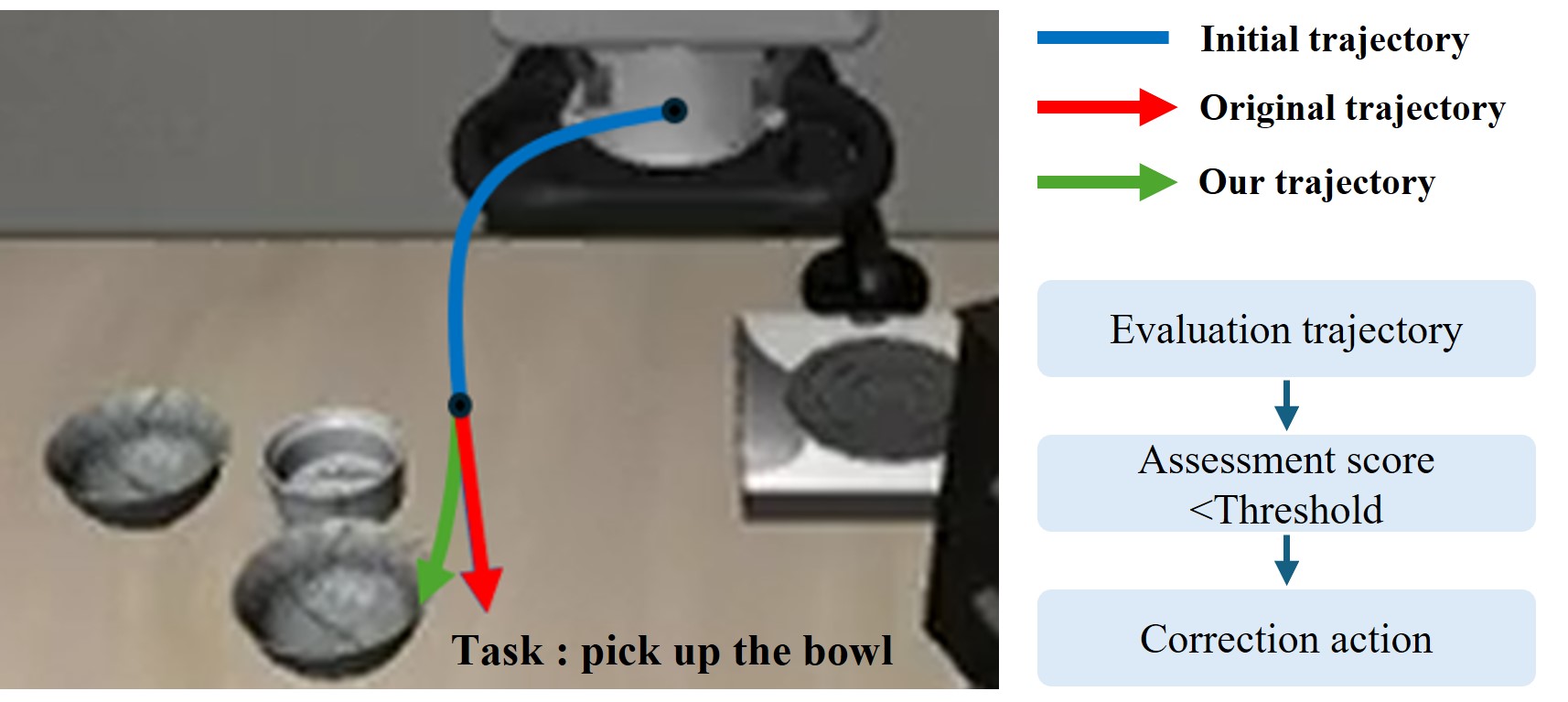}
        \caption{Self-correction of actions for accuracy issues}
        \label{fig:sub_a}
    \end{subfigure}

\vspace{0.2cm}
    \begin{subfigure}{\linewidth} 
        \centering
        \includegraphics[width=0.9\linewidth]{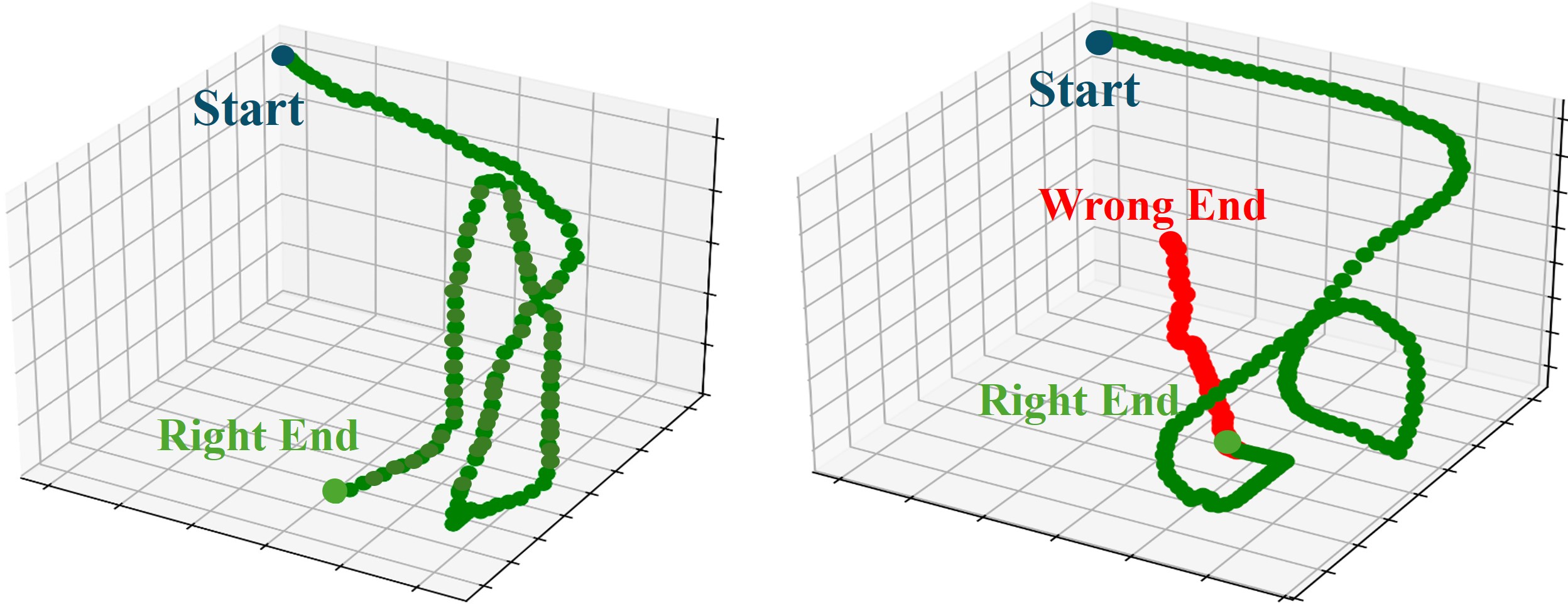}
        \caption{Visualization of superfluous actions due to failed termination.}
        \label{fig:sub_b}
    \end{subfigure}

    \caption{The VLA-SCT framework addresses two fundamental challenges in VLAs. (a) For motion accuracy, it corrects a deviating trajectory. (b) For task termination, it prevents superfluous actions after task completion.}
    \label{fig:main}
    \vspace{-0.5cm}
\end{figure}
\noindent  deficiency often stems from the predominantly open-loop nature of their execution policies, which lack the real-time feedback mechanisms necessary to correct minor deviations during task execution. Consequently, tasks demanding high-precision control, such as dexterous grasping or placement, are prone to trajectory deviations, unstable grasps, and outright execution failures\cite{PertschK-RSS-25}. This limitation is particularly pronounced in applications such as medical and industrial automation, where sub-millimeter accuracy is often a prerequisite\cite{gu2025humanoid}. As illustrated in Figure \ref{fig:sub_a}, the open-loop nature of baseline VLAs can result in execution deviations, such as missing the target bowl (red line). Our proposed self-correction mechanism addresses this issue by first evaluating the trajectory and then applying a corrective action (green line) to ensure precise task completion.
\begin{figure*}[t]
    \centering 
    \includegraphics[width=0.9\textwidth]{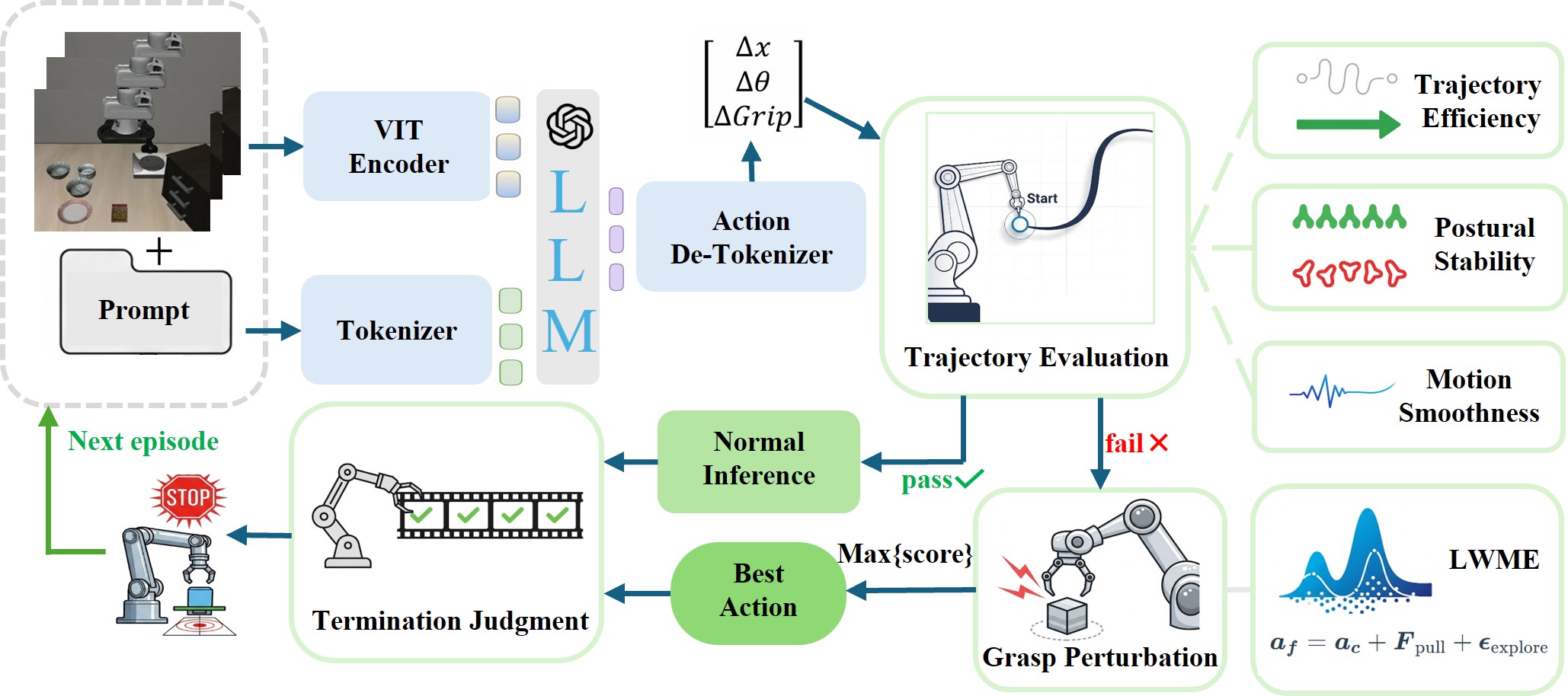} 
    \caption{ An overview of the VLA-SCT framework. Our method enhances a standard VLA model by introducing a conditional control loop for action self-correction and a visual judgment module for reliable task termination.} 
    \label{fig:zhutu} 
\end{figure*}
Second, a prevalent limitation is the absence of a robust task termination mechanism, causing many VLA models to fail in recognizing when a given goal has been accomplished. Consequently, these models often exhibit non-terminating behavior, continuing to execute superfluous actions until a predefined step limit is reached, which results in the task being erroneously classified as a failure\cite{li-etal-2025-rate}. Such behavior severely undermines both operational efficiency and system reliability, posing a significant safety risk and a fundamental challenge to the deployment of autonomous agents in open-world settings\cite{jia-etal-2025-hierarchical,wei2024physical}. Figure \ref{fig:sub_b} provides a clear illustration of this failure mode, where the robotic arm, despite having successfully completed its task, continues its motion until it exceeds the maximum step count, leading to a timeout failure.

To address these issues, we propose a general \textbf{S}elf-\textbf{C}orrection and \textbf{T}ermination framework, the \textbf{VLA-SCT} framework, which systematically improves the robustness and reliability of the VLA model by integrating self-correction and termination detection mechanisms. Our main contributions are summarized as follows:
\begin{itemize}[leftmargin=*]
\item We propose and implement a general, lightweight VLA-SCT framework designed to enhance existing VLA models. By building a modular intelligent control layer external to the VLA model, the framework reduces the risk of failures in both action execution and mission termination.
\item We propose two data-driven mechanisms: an online self-calibration module based on Locally Weighted Moment Estimation(LWME)
and a nonparametric termination decision module based on visual feature matching.
\item Extensive experiments show that our VLA-SCT framework not only achieved the highest average success rate, but also significantly improved the inference efficiency, demonstrating the efficiency and great potential of the framework.
\end{itemize}
\vspace{-0.2cm}
\section{Methods}
\label{sec:majhead}

The VLA-SCT framework is a modular intelligent control layer that enhances existing VLA models, as illustrated in Figure \ref{fig:zhutu}. This framework introduces an evaluation-correction-termination feedback loop that improves execution robustness by processing the baseline model's initial motion plans. The following subsections provide a detailed explanation of the three core components of this loop: the Trajectory Evaluation module,  Grasp Perturbation module, and Termination Detection module.
\vspace{-0.3cm}
\subsection{Trajectory Evaluation}
\label{ssec:subhead}

At the initial stage of a mission, we designed a lightweight trajectory quality evaluator that analyzes the efficiency, smoothness, and stability of the robot's early motion trajectories to predict the risk of mission failure.\\
\textbf{Trajectory Efficiency. }We define trajectory efficiency $S_{eff}$ as a normalized score that is inversely proportional to the total curvature and total torsion of the path, which is defined as
\begin{equation}
S_{eff}=\frac{1}{1+a\int_{\gamma}\kappa(s)ds+b\int_{\gamma}|\tau(s)|ds}\times100\%,
\end{equation}
where $\gamma$ represents a segment of the three-dimensional spatial trajectory of the robot end effector within the evaluation window, $a,b$ are parameters,
$s$ is the arc length parameter of the trajectory,
$\kappa(s)$ is the curvature of the trajectory at point $s$, 
$\tau(s)$ is the torsion of the trajectory at point $s$.\\
\textbf{Postural Stability. }To evaluate trajectory posture stability, we adopt methods from differential geometry and model each robot posture as a point on the SO(3) manifold. Our posture stability score is defined as a function that decays exponentially with the accumulated geodesic distance along the trajectory, which is defined as
\begin{equation}S_{sta}=\exp\left(-k\sum_{t}\arccos\left(\frac{\mathrm{Tr}(\mathbf{R_tR_{t-1}^T)}-1}{2}\right)\right),\end{equation}
where $\mathbf{R_t}$ is a 3x3 rotation matrix representing the three-dimensional spatial orientation of the robot's end effector at discrete time step $t$, and $k$ is the smoothing coefficient. This metric accurately reflects the temporal continuity of the trajectory by accumulating the shortest rotation angle between each pair of consecutive orientations.\\
\textbf{Motion Smoothness. }Our evaluation metric is based on quantifying a trajectory's total jerk. This approach follows the Minimum Jerk Principle, a gold standard in biomechanics and robotics for describing smooth, natural motion as one that minimizes the rate of change of acceleration. Given a robot trajectory $\mathcal{T}=\{(\bm{p_1},\bm{q_1}),\ldots,(\bm{p_N},\bm{q_N)}\}$ consisting of $N$ pose points, where $\bm{p_i}$ is the position vector,
$\bm{q_i}$ is the quaternion representing orientation, and the sampling time interval is $\Delta t$, the score is defined as
\begin{equation}S_{smo}=\exp(-\mu\sum_i\left(\|\bm{j_{p,i}}\|^2+w_{rot}\cdot\|\bm{\zeta_{q,i}}\|^2\right)\Delta t),\end{equation}
where $\bm{j_{p,i}}$ is the translational jerk, $\bm{\zeta_{q,i}}$ is the angular jerk, $\mu$ is the sensitivity coefficient, and $w_\mathrm{rot}$ is the rotational weight.
\vspace{-0.3cm}
\subsection{Grasp Perturbation}
\label{ssec:subhead}
When the Trajectory Evaluation module determines that the initial action plan is low quality, the VLA-SCT framework activates the Grasp Perturbation module. This module does not simply replace the original action but operates as an experience-based online adaptive correction mechanism. Its core principle is to generate targeted perturbations using the probability distribution dynamically constructed from the current visual environment—this distribution consists of historically successful actions—to guide potentially unsuccessful actions toward areas with higher success probability. We use the radial basis function (RBF) kernel to measure the similarity between the current visual feature vector $\bm{v_c}$ and the visual features $\bm{v_i}$ of each successful historical experience in the memory bank. This similarity value is used as the weight $w_i$:
\begin{equation}
w_i=\exp(-\gamma\|\bm{v_c}-\bm{v_i}\|^2),
\end{equation}
where $\gamma$ is the hyperparameter that controls the decay rate. The weight $w_i$ indicates the relevance of the i-th historical successful action, $\bm{a_i}$, to the current situation. Using the calculated weights $w_i$, we perform weighted statistics on the relevant historical successful actions $a_i$ to estimate the center and shape of the successful action distribution. First, we calculate the weighted average $\bm{a_\mu}$, which represents the density center of the successful action distribution in this visual scene:
\begin{equation}\bm{a_{\mu}}=\dfrac{\sum_iw_i\bm{a_i}}{\sum_iw_i}.\end{equation}
Next, we calculate the weighted covariance matrix, which describes the distribution shape, direction, and correlation between dimensions of successful actions in the seven-dimensional action space:
\begin{equation}\mathbf{Q}=\dfrac{\sum_iw_i(\bm{a_i}-\bm{a_{\mu}})(\bm{a_i}-\bm{a_{\mu}})^T}{\sum_iw_i}.\end{equation}
The covariance matrix $\mathbf{Q}$ is key to generating structured random perturbations. To ensure numerical stability, we regularize it to $\mathbf{Q_{reg}} =\mathbf{Q} + \lambda \mathbf{I}$, where $\lambda$ is a minimal constant to avoid matrix singularities. The final perturbation we designed consists of three parts: a deterministic gravity term that pulls the action towards the success region, an anisotropic noise term that explores the distribution pattern of success, and an isotropic noise term that promotes generalization, as follows:
\begin{equation}
\bm{a_f}=\bm{a_c}+\alpha(\bm{a_\mu}-\bm{a_c})+\sqrt{T}\mathcal{N}(0,\beta \mathbf{Q_\text{reg}})+\mathcal{N}(0,\sigma^2\mathbf{I}),\end{equation}
where $\alpha$ and $\beta$ are controllable parameters, and $T$ is the sampling temperature. At the implementation level, in addition to regularizing the covariance matrix, we also need to clip the resulting motion vectors (such as the gripper opening and closing amplitude) to ensure that all output values are within the robot's physical limits, thereby ensuring the algorithm's stability and safety.

\vspace{-0.3cm}
\subsection{Termination Detection}
\label{ssec:subhead}
Our termination detection module determines task completion by comparing the current visual state against a dynamically updated repository $\mathcal{M}=\{\bm{I_1},\bm{I_2},…,\bm{I_N}\}$, where $\bm{I_i}$ represents successful visual states from the past. This non-parametric approach avoids hard-coded rules, making it highly generalizable and adaptive. At each step, the module compares the current camera image $\bm{I_0}$ with each image $\bm{I_i}$ in the repository. 
We preprocess and flatten both images into one-dimensional vectors, $\bm{X}=\{x_1,x_2,\dots,x_n\}$ and $\bm{Y}=\{y_1,y_2,\dots,y_n\}$, where $n$ is the total number of pixels. We use the Pearson correlation coefficient $r$ as the core similarity metric because it is robust to changes in brightness and contrast. Instead of a direct pixel-wise comparison, $r$ is calculated as the standardized covariance between the two vectors, obtained by summing the product of the centered deviations $(x_i - \mu_x)$ and $(y_i - \mu_y)$ across all pixels, normalized by the product of the vectors' standard deviations, where $\mu_x$ and $\mu_y$ denote the mean of the elements in vectors $\bm{X}$ and $\bm{Y}$. This coefficient is then linearly converted to an intuitive similarity score $S=\frac{r+1}{2}$, ranging from 0 to 1. The module finds the maximum similarity score $S_\text{max}$ between the current view and all stored memories. If this score exceeds a preset termination threshold $\tau$, the task is considered complete and a stop signal is issued. Otherwise, execution continues. The threshold $\tau$ is a tunable hyperparameter that allows us to adjust the system's termination strategy from conservative to aggressive.

\vspace{-0.0cm}
\section{Experiments}
\label{ssec:subhead}
\vspace{-0.0cm}
\subsection{Dataset and Metrics}
\label{ssec:subhead}

All of our experiments were conducted on the LIBERO benchmark\cite{liu2023libero}. LIBERO is a widely used benchmark in the fields of VLA and robot learning. We evaluated the model based on task success rate and inference speed. Task success rate is the core metric for evaluating the framework's task execution capabilities. It directly reflects the actual effectiveness and robustness of our proposed self-correction and termination module in complex operational scenarios.
\vspace{-0.6cm}
\subsection{Implementation Details}
\label{ssec:subhead}

Our experiments use the publicly available OpenVLA-7B\cite{kim2025openvla} model as the baseline.  Our proposed VLA-SCT framework is an inference-only augmentation module that requires no training. This design aligns with recent efficient approaches in other domains, such as speech recognition error correction\cite{yang2023generative}. Therefore, the pre-trained OpenVLA model weights remain frozen in all experiments. All input images are processed to the OpenVLA standard of 224×224 pixels. Experiments are conducted on a single NVIDIA RTX 4090 GPU, with the model loaded at bfloat16 precision, requiring approximately 15GB of CUDA memory.

\begin{table*}[t!]
    \centering
    \renewcommand{\arraystretch}{1.0} 
    \caption{Performance comparison of our proposed method with state-of-the-art methods on the LIBERO benchmark, where "+R" denotes the preservation
 of relative token positions, "+S" represents the incorporation of Canny edge information.}
    \label{tab:my_comparison}
    \begin{tabular*}{\textwidth}{l @{\extracolsep{\fill}} ccccc}
        \toprule
        & \multicolumn{5}{c}{\textbf{Success Rate (\%, $\uparrow$) / Speed up ($\uparrow$)}} \\
        \cmidrule(r){2-6}
        \textbf{Method} & \textbf{Goal} & \textbf{Object} & \textbf{Spatial} & \textbf{Long} & \textbf{Average} \\
        \midrule
        OpenVLA \cite{kim2025openvla}      & 76.40 / 1.00 & 88.80 / 1.00 & 83.20 / 1.00 & 53.40 / 1.00 & 75.45 / 1.00 \\
        \midrule
        SparseVLM \cite{zhangsparsevlm}    & 74.20 / 1.33 & 84.00 / 1.33 & 83.40 / 1.33 & 52.80 / 1.33 & 73.60 / 1.33 \\
        FoPru + R \cite{jiang2024fopru}        & 59.80 / 1.29 & 81.20 / 1.29 & 71.60 / 1.30 & 26.20 / 1.35 & 59.70 / 1.31 \\
        PruMerge + R \cite{shang2024llava} & 0.00 / 1.54  & 0.00 / 1.32  & 0.00 / 1.27  & 0.00 / 1.32  & 0.00 / 1.36  \\
        FastVLM + R + S \cite{vasu2025fastvlm} & 73.20 / 1.21 & 77.00 / 1.11 & 79.80 / 1.12 & 36.60 / 1.20 & 66.65 / 1.16 \\
        VisionZip + R + S \cite{yang2025visionzip} & 46.00 / 1.20 & 47.40 / 1.23 & 34.20 / 1.19 & 4.60 / 1.22 & 33.05 / 1.21 \\
        SP-VLA
        \cite{li2025spvla} & 75.40 / 1.46 & 85.60 / 1.30 & 84.40 / 1.30 & 
 54.20 / 1.32 & 74.90 / 1.35 \\
        \midrule
        \textbf{VLA-SCT(Ours)}        & \textbf{82.00} / 1.22 & \textbf{92.80} / 1.09 & \textbf{91.20} / 1.10 & \textbf{60.20} / 1.06 & \textbf{81.55} / 1.12 \\
        \bottomrule
    \end{tabular*}
\end{table*}
\begin{table}[t]
    \centering 
    \renewcommand{\arraystretch}{1.0} 
    \caption{Ablation study on the effectiveness of individual modules. Module A is Trajectory Evaluation; Module B is Grasp Perturbation; Module C is Termination Detection.}
    \label{tab:ablation_modified}
    
    \setlength\tabcolsep{3pt}

    \begin{tabularx}{\columnwidth}{lCCCcc}
        \toprule
        \multirow{2}{*}{\textbf{Model}} & \multicolumn{2}{c}{\textbf{Modules}}  & \multirow{2}{*}{\textbf{Success Rate (\%)}}\\
        \cmidrule(lr){2-3} 
         & A+B & C& & \\
        \midrule
        \textbf{VLA-SCT}  & \checkmark   & \checkmark   & \textbf{91.20} \\
        -         & \checkmark   & $\times$     & 87.40 \\
        -    & $\times$   & \checkmark    & 85.60 \\
        baseline     & $\times$   & $\times$   & 83.20 \\
        \bottomrule
    \end{tabularx}
\end{table}
\vspace{-0.3cm}
\subsection{Experimental Results}
\label{ssec:subhead}
\begin{figure}[t]
    \centering
    \includegraphics[width=1.0\linewidth]{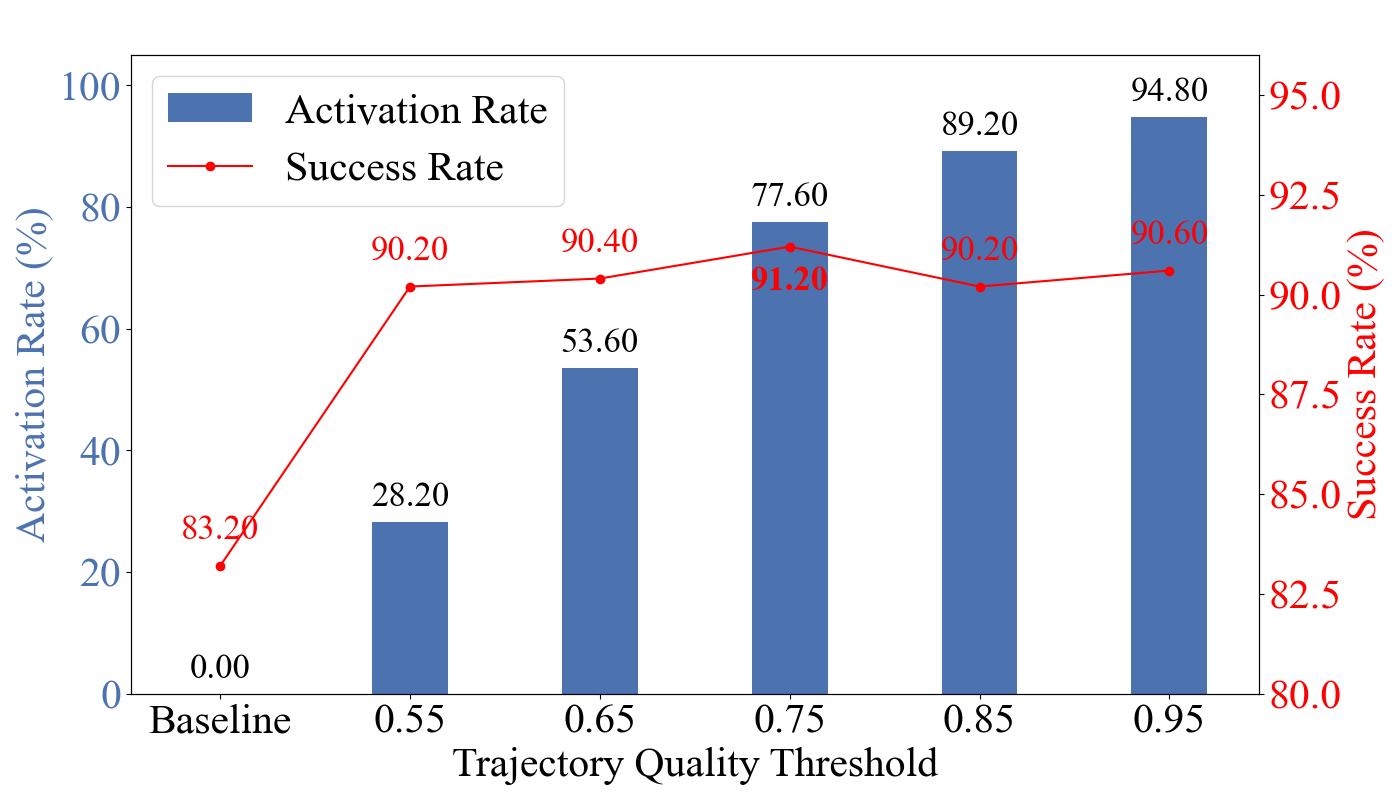}
    \caption{Sensitivity analysis of the trajectory quality threshold. }
    \label{sense}
\end{figure}

\textbf{Comparison with other method.}
As presented in Table \ref{tab:my_comparison}, our comprehensive experimental results clearly demonstrate that the proposed VLA-SCT framework establishes a new state-of-the-art on the challenging LIBERO benchmark. Our method achieves an average success rate of 81.55\%, representing an absolute improvement of 6.1\% over the OpenVLA baseline (75.45\%) and surpassing all compared approaches. This advantage holds consistently across all four task categories, with VLA-SCT attaining the best performance in Goal (82.00\%), Object (92.80\%), Spatial (91.20\%), and even in the challenging Long tasks (60.20\%). Notably, VLA-SCT achieves high task success without sacrificing efficiency. While SP-VLA attains greater speedup at the cost of accuracy, our approach delivers the best overall trade-off, combining state-of-the-art success rates with a 1.12× speedup over OpenVLA. This shows that the VLA-SCT framework not only corrects erroneous actions but also enables the model to terminate in time after completing the task. This unique combination of superior accuracy and practical efficiency validates that our framework gains its advantage by enhancing the model's intrinsic decision-making quality, rather than sacrificing performance for speed, highlighting its robustness and value for real-world robotic applications.\\
\textbf{Ablation Study. }To quantify the contribution of each core component, we conducted an ablation study on the LIBERO spatial suite, as summarized in Table \ref{tab:ablation_modified}. While the holistic efficacy of our framework is established in Table \ref{fig:main}, this subset is selected for its stringent accuracy requirements, enabling a granular analysis of individual module contributions. Our analysis begins with the baseline model, which achieves a success rate of 83.20\% without any of our proposed modules. Incorporating solely the self-correction mechanism (Modules A+B) boosts the success rate to 87.40\%, validating the efficacy of the Trajectory Evaluation and Grasp Perturbation modules in enhancing manipulation precision. Likewise, deploying Module C alone boosts success to 85.60\% by mitigating timeout-induced failures. Crucially, when all components are combined in our full VLA-SCT framework, the model achieves the highest success rate of 91.20\%, confirming that the self-correction and termination mechanisms provide substantial and complementary benefits. This study validates that both the enhancement of action precision and the reliable identification of task completion are integral to the framework's overall state-of-the-art performance.\\
\textbf{Sensitivity Analysis. }To investigate the impact of key hyperparameters within our framework, we performed a sensitivity analysis on the trajectory quality threshold, as illustrated in Figure \ref{sense}. This threshold functions as the core gatekeeper that determines the activation of our  Grasp Perturbation module. Stricter thresholds drive a monotonic rise in activation (28.20\% at 0.55 to 94.80\% at 0.95), consistent with the expectation of increased interventions.
\noindent  Notably, the success rate exhibits a distinct peak. The success rate reaches its maximum of 91.20\% at a threshold of 0.75. However, as the threshold is further increased to 0.85 and 0.95, the success rate slightly declines despite the continually rising activation rate. This trend highlights a trade-off: an excessively permissive threshold fails to capture all errors, whereas an overly stringent one leads to over-correction, where superfluous interventions on valid trajectories may introduce instability. Thus, we identify 0.75 as optimal, balancing success maximization against superfluous interventions.
\vspace{-0.2cm}
\section{CONCLUSION}
This paper proposes VLA-SCT, an innovative general framework for enhancing pre-trained VLA models. We summarize the limitations of existing VLA models in fine manipulation and reliable task completion, and improve manipulation accuracy and task reliability by integrating Trajectory Evaluation, Grasp Perturbation, and Termination Detection modules. Comparative and ablation studies on LIBERO validate our SOTA performance, advancing reliable VLA autonomy. A key direction for future work is investigating the applicability of our framework to diverse VLA architectures to improve cross-model generalization.

\label{ssec:subhead}
\clearpage

{
\bibliographystyle{IEEEbib}
\bibliography{strings,refs}

\begin{thebibliography}{10}

\bibitem{gu2025humanoid}
Zhaoyuan Gu, Junheng Li, Wenlan Shen, Wenhao Yu, Zhaoming Xie, Stephen McCrory, Xianyi Cheng, Abdulaziz Shamsah, Robert Griffin, C~Karen Liu, et~al.,
\newblock ``Humanoid locomotion and manipulation: Current progress and challenges in control, planning, and learning,''
\newblock {\em IEEEASME transactions on mechatronics}, 2025.

\bibitem{liurdt}
Songming Liu, Lingxuan Wu, Bangguo Li, Hengkai Tan, Huayu Chen, Zhengyi Wang, Ke~Xu, Hang Su, and Jun Zhu,
\newblock ``Rdt-1b: a diffusion foundation model for bimanual manipulation,''
\newblock in {\em The Thirteenth International Conference on Learning Representations}, 2025.

\bibitem{11127486}
Ziqi Jia, Junjie Li, Xiaoyang Qu, and Jianzong Wang,
\newblock ``Enhancing multi-agent systems via reinforcement learning with llm-based planner and graph-based policy,''
\newblock in {\em 2025 IEEE International Conference on Robotics and Automation (ICRA)}, 2025, pp. 1240--1246.

\bibitem{brohan2022rt}
Anthony Brohan, Noah Brown, Justice Carbajal, Yevgen Chebotar, Joseph Dabis, Chelsea Finn, Keerthana Gopalakrishnan, Karol Hausman, Alexander Herzog, Jasmine Hsu, et~al.,
\newblock ``Rt-1: Robotics transformer for real-world control at scale,''
\newblock {\em Robotics: Science and Systems XIX}, 2023.

\bibitem{zitkovich2023rt}
Brianna Zitkovich, Tianhe Yu, Sichun Xu, Peng Xu, Ted Xiao, Fei Xia, Jialin Wu, Paul Wohlhart, Stefan Welker, Ayzaan Wahid, et~al.,
\newblock ``Rt-2: Vision-language-action models transfer web knowledge to robotic control,''
\newblock in {\em Conference on Robot Learning}. PMLR, 2023, pp. 2165--2183.

\bibitem{o2024rt-x}
Abby O’Neill, Abdul Rehman, Abhiram Maddukuri, Abhishek Gupta, Abhishek Padalkar, Abraham Lee, Acorn Pooley, Agrim Gupta, Ajay Mandlekar, Ajinkya Jain, et~al.,
\newblock ``Open x-embodiment: Robotic learning datasets and rt-x models: Open x-embodiment collaboration,''
\newblock in {\em 2024 IEEE International Conference on Robotics and Automation (ICRA)}. IEEE, 2024, pp. 6892--6903.

\bibitem{kim2025openvla}
Moo~Jin Kim, Karl Pertsch, Siddharth Karamcheti, Ted Xiao, Ashwin Balakrishna, Suraj Nair, Rafael Rafailov, Ethan~P Foster, Pannag~R Sanketi, Quan Vuong, et~al.,
\newblock ``Openvla: An open-source vision-language-action model,''
\newblock in {\em Conference on Robot Learning}. PMLR, 2025, pp. 2679--2713.

\bibitem{black2410pi0}
Kevin Black, Noah Brown, Danny Driess, Adnan Esmail, Michael Equi, Chelsea Finn, Niccolo Fusai, Lachy Groom, Karol Hausman, Brian Ichter, et~al.,
\newblock ``$\pi_0$: A vision-language-action flow model for general robot control,''
\newblock {\em arXiv preprint arXiv.2410.24164}, 2024.

\bibitem{black2025pi0.5}
Kevin Black, Noah Brown, James Darpinian, Karan Dhabalia, Danny Driess, Adnan Esmail, Michael Equi, Chelsea Finn, Niccolo Fusai, Manuel~Y Galliker, et~al.,
\newblock ``$\pi_{0.5}$ a vision-language-action model with open-world generalization,''
\newblock {\em arXiv preprint arXiv:2504.16054}, 2025.

\bibitem{zhangsparsevlm}
Yuan Zhang, Chun-Kai Fan, Junpeng Ma, Wenzhao Zheng, Tao Huang, Kuan Cheng, Denis~A Gudovskiy, Tomoyuki Okuno, Yohei Nakata, Kurt Keutzer, et~al.,
\newblock ``Sparsevlm: Visual token sparsification for efficient vision-language model inference,''
\newblock in {\em Forty-second International Conference on Machine Learning}, 2025.

\bibitem{vasu2025fastvlm}
Pavan Kumar~Anasosalu Vasu, Fartash Faghri, Chun-Liang Li, Cem Koc, Nate True, Albert Antony, Gokula Santhanam, James Gabriel, Peter Grasch, Oncel Tuzel, et~al.,
\newblock ``Fastvlm: Efficient vision encoding for vision language models,''
\newblock in {\em Proceedings of the Computer Vision and Pattern Recognition Conference}, 2025, pp. 19769--19780.

\bibitem{zhong2025survey}
Yifan Zhong, Fengshuo Bai, Shaofei Cai, Xuchuan Huang, Zhang Chen, Xiaowei Zhang, Yuanfei Wang, Shaoyang Guo, Tianrui Guan, Ka~Nam Lui, et~al.,
\newblock ``A survey on vision-language-action models: An action tokenization perspective,''
\newblock {\em arXiv preprint arXiv:2507.01925}, 2025.

\bibitem{PertschK-RSS-25}
Karl Pertsch, Kyle Stachowicz, Brian Ichter, Danny Driess, Suraj Nair, Quan Vuong, Oier Mees, Chelsea Finn, and Sergey Levine,
\newblock ``{FAST: Efficient Action Tokenization for Vision-Language-Action Models},''
\newblock in {\em Proceedings of Robotics: Science and Systems}, LosAngeles, CA, USA, June 2025.

\bibitem{li-etal-2025-rate}
Junjie Li, Nan Zhang, Xiaoyang Qu, Kai Lu, Guokuan Li, Jiguang Wan, and Jianzong Wang,
\newblock ``{RATE}-nav: Region-aware termination enhancement for zero-shot object navigation with vision-language models,''
\newblock in {\em Findings of the Association for Computational Linguistics: ACL 2025}, 2025, pp. 6564--6574.

\bibitem{jia-etal-2025-hierarchical}
Ziqi Jia, Anmin Wang, Xiaoyang Qu, Xiaowen Yang, and Jianzong Wang,
\newblock ``Hierarchical-task-aware multi-modal mixture of incremental {L}o{RA} experts for embodied continual learning,''
\newblock in {\em Proceedings of the 63rd Annual Meeting of the Association for Computational Linguistics}, 2025, pp. 28415--28427.

\bibitem{wei2024physical}
Hui Wei, Hao Tang, Xuemei Jia, Zhixiang Wang, Hanxun Yu, Zhubo Li, Shin’ichi Satoh, Luc Van~Gool, and Zheng Wang,
\newblock ``Physical adversarial attack meets computer vision: A decade survey,''
\newblock {\em IEEE Transactions on Pattern Analysis and Machine Intelligence}, vol. 46, no. 12, pp. 9797--9817, 2024.

\bibitem{liu2023libero}
Bo~Liu, Yifeng Zhu, Chongkai Gao, Yihao Feng, Qiang Liu, Yuke Zhu, and Peter Stone,
\newblock ``Libero: Benchmarking knowledge transfer for lifelong robot learning,''
\newblock {\em Advances in Neural Information Processing Systems}, vol. 36, pp. 44776--44791, 2023.

\bibitem{yang2023generative}
Chao-Han~Huck Yang, Yile Gu, Yi-Chieh Liu, Shalini Ghosh, Ivan Bulyko, and Andreas Stolcke,
\newblock ``Generative speech recognition error correction with large language models and task-activating prompting,''
\newblock in {\em 2023 IEEE Automatic Speech Recognition and Understanding Workshop (ASRU)}. IEEE, 2023, pp. 1--8.

\bibitem{jiang2024fopru}
Lei Jiang, Weizhe Huang, Tongxuan Liu, Yuting Zeng, Jing Li, Lechao Cheng, and Xiaohua Xu,
\newblock ``Fopru: Focal pruning for efficient large vision-language models,''
\newblock {\em arXiv preprint arXiv:2411.14164}, 2024.

\bibitem{shang2024llava}
Yuzhang Shang, Mu~Cai, Bingxin Xu, Yong~Jae Lee, and Yan Yan,
\newblock ``Llava-prumerge: Adaptive token reduction for efficient large multimodal models,''
\newblock in {\em Proceedings of the IEEE/CVF International Conference on Computer Vision}, 2025, pp. 22857--22867.

\bibitem{yang2025visionzip}
Senqiao Yang, Yukang Chen, Zhuotao Tian, Chengyao Wang, Jingyao Li, Bei Yu, and Jiaya Jia,
\newblock ``Visionzip: Longer is better but not necessary in vision language models,''
\newblock in {\em Proceedings of the Computer Vision and Pattern Recognition Conference}, 2025, pp. 19792--19802.

\bibitem{li2025spvla}
Ye~Li, Yuan Meng, Zewen Sun, Kangye Ji, Chen Tang, Jiajun Fan, Xinzhu Ma, Shutao Xia, Zhi Wang, and Wenwu Zhu,
\newblock ``Sp-vla: A joint model scheduling and token pruning approach for vla model acceleration,''
\newblock {\em arXiv preprint arXiv:2506.12723}, 2025.

\end{thebibliography}
}

\end{document}